\documentclass[sigconf]{acmart}
\usepackage{bm}
\def\method{RAHNet}
\def\methodcl{BSCL}

\usepackage{multirow}
\usepackage{enumitem}
\usepackage{tabularx}
\newcolumntype{Y}{>{\small\centering\arraybackslash}X}
\usepackage{graphicx}
\usepackage{subfigure}
\usepackage{stfloats}
\usepackage{algorithmic}
\usepackage[ruled]{algorithm2e}
\usepackage{bbding}
\usepackage{amsfonts}
\usepackage{colortbl}
\usepackage{amsmath}

\definecolor{LightCyan}{rgb}{0.88,1,1}

\newcommand{\ie}{\textit{i}.\textit{e}.}

\AtBeginDocument{%
  \providecommand\BibTeX{{%
    \normalfont B\kern-0.5em{\scshape i\kern-0.25em b}\kern-0.8em\TeX}}}

\copyrightyear{2023}
\acmYear{2023}
\setcopyright{acmlicensed}\acmConference[MM '23]{Proceedings of the 31st
ACM International Conference on Multimedia}{October 29-November 3,
2023}{Ottawa, ON, Canada}
\acmBooktitle{Proceedings of the 31st ACM International Conference on
Multimedia (MM '23), October 29-November 3, 2023, Ottawa, ON, Canada}
\acmPrice{15.00}
\acmDOI{10.1145/3581783.3612360}
\acmISBN{979-8-4007-0108-5/23/10}

\begin{document}

\title[\method{}]{\method{}: Retrieval Augmented Hybrid Network for \\Long-tailed Graph Classification}


% \def\affil{National Key Laboratory for Multimedia Information Processing, School of Computer Science, Peking University}
\def\affil{National Key Laboratory for Multimedia Information Processing, \\School of Computer Science, \\ Peking University}

\author{Zhengyang Mao}
\authornote{Both authors contributed equally to this research.}
\affiliation{%
  \institution{\affil{}}
  % \streetaddress{}
  \city{Beijing}
  \country{China}
}
\email{zhengyang.mao@stu.pku.edu.cn}

\author{Wei Ju}
\authornotemark[1]
\affiliation{%
  \institution{\affil{}}
  % \streetaddress{}
  \city{Beijing}
  \country{China}
}
\email{juwei@pku.edu.cn}
\orcid{0000-0001-9657-951X}

\author{Yifang Qin}
\affiliation{%
  \institution{\affil{}}
  % \streetaddress{}
  \city{Beijing}
  \country{China}
}
\email{qinyifang@pku.edu.cn}

\author{Xiao Luo}
\authornote{Corresponding authors.}
\orcid{0000-0002-7987-3714}
\affiliation{%
\institution{Department of Computer Science, University of California, Los Angeles}
%   \streetaddress{}
  \city{Los Angeles}
  \country{USA}}
\email{xiaoluo@cs.ucla.edu}

\author{Ming Zhang}
% \authornote{Corresponding authors.}
\authornotemark[2]
\orcid{0000-0002-9809-3430}
\affiliation{%
  \institution{\affil{}}
  % \streetaddress{}
  \city{Beijing}
  \country{China}
}
\email{mzhang_cs@pku.edu.cn}

\renewcommand{\shortauthors}{Zhengyang Mao, Wei Ju, Yifang Qin, Xiao Luo, \& Ming Zhang}


% \ccsdesc[500]{Mathematics of computing~Graph algorithms}


\begin{abstract}

Graph classification is a crucial task in many real-world multimedia applications, where graphs can represent various multimedia data types such as images, videos, and social networks. Previous efforts have applied graph neural networks (GNNs) in balanced situations where the class distribution is balanced. However, real-world data typically exhibit long-tailed class distributions, resulting in a bias towards the head classes when using GNNs and limited generalization ability over the tail classes. Recent approaches mainly focus on re-balancing different classes during model training, which fails to explicitly introduce new knowledge and sacrifices the performance of the head classes. To address these drawbacks, we propose a novel framework called Retrieval Augmented Hybrid Network (\method{}) to jointly learn a robust feature extractor and an unbiased classifier in a decoupled manner. In the feature extractor training stage, we develop a graph retrieval module to search for relevant graphs that directly enrich the intra-class diversity for the tail classes. Moreover, we innovatively optimize a category-centered supervised contrastive loss to obtain discriminative representations, which is more suitable for long-tailed scenarios. In the classifier fine-tuning stage, we balance the classifier weights with two weight regularization techniques, \ie, Max-norm and weight decay. Experiments on various popular benchmarks verify the superiority of the proposed method against state-of-the-art approaches.

\end{abstract}

\maketitle

\section{Introduction}

Graph classification, which captures the graph-level properties to predict classes for graphs, is a fundamental task in data mining with many real-world multimedia applications across various domains ~\cite{lei2020novel,peng2021attention}. For example, social networks \cite{zhang2022improving}, images \cite{dong2022weighted}, and knowledge graphs \cite{cao2022cross} can all be represented as graphs, and graph classification can be used in multimedia tasks such as object recognition, semantic segmentation, and recommender systems. As a powerful technique for graph representation learning, graph neural networks (GNNs) ~\cite{kipf2017semi,ahmed2017inductive,velivckovic2017graph,ju2023comprehensive} have achieved remarkable success in various applications, such as graph classification ~\cite{ju2022kgnn,ju2023tgnn,luo2022dualgraph,luo2023towards}, novel drug discovery ~\cite{ma2021gf,rathi2019practical,bongini2021molecular,ju2023few}, traffic forecasting~\cite{li2021spatial,fang2021spatial,zhao2023dynamic}, and recommender systems ~\cite{wei2019mmgcn,wei2021contrastive,ju2022kernel,qin2023disenpoi,qin2023learning}. GNNs are designed to propagate and aggregate messages on a graph, where each node obtains messages from all of its neighbors and then performs neighborhood aggregation and representation combination iteratively. Finally, node representations can be integrated into graph representations by pooling operations, and thus both the structural information and attributive knowledge of individual nodes can be implicitly merged into the graph-level representation \cite{gilmer2017neural}.

Although existing GNNs have achieved excellent performance, these methods typically concentrate on a balanced data-split setting. However, realistic data tends to follow the long-tailed class distribution ~\cite{peng2021long,wang2022balanced}. It is primarily evident from the fact that a minority of dominant classes (i.e. head classes) usually occupy a large amount of data, while the majority of classes (i.e. tail classes) each contains very few data samples. In such data distributions, naive GNNs often degrade and result in sub-optimal classification performance on tail classes due to two major limitations: (i) models learned from the long-tailed distribution can be easily biased towards head classes \cite{zhang2021deep}, and (ii) models fail to generalize well for tail classes because of insufficient training data. Undoubtedly, the long-tailed characteristics immensely restrict the practical use of GNNs, hence it is essential to develop tailored GNN methods for realistic long-tailed graph data.

Recently, long-tailed learning has received widespread attention in the context of neural networks ~\cite{liu2022long,cai2022luna,zhu2022balanced,li2022nested,li2022trustworthy,wang2020long,zhang2022self}, there exist three active strands of work: class re-balancing, information augmentation, and decoupling training. Class re-balancing attempts to balance the training data of different classes during model training, either by re-sampling the data ~\cite{jiang2021improving,guo2021long} or using cost-sensitive learning ~\cite{park2021influence,cao2019learning,wang2017learning}. Information augmentation, on the other hand, seeks to improve model performance by introducing additional knowledge through a transfer learning approach ~\cite{park2021influence,parisot2022long}. Moreover, recent work has shown that decoupling feature learning and classifier learning can lead to improved performance compared to traditional end-to-end training methods ~\cite{kang2020decoupling}. While there are several long-tailed algorithms designed for node-level classification on graphs ~\cite{liu2021tail,song2022tam,park2021graphens,yun2022lte4g}, the problem of long-tailed recognition for graph-level classification is still largely unexplored and challenging, which is the main focus of this paper.

Despite achieving encouraging performance, most of the existing algorithms still suffer from two key limitations. Firstly, most class re-balancing approaches are inclined to improve the performance of tail classes, but this often comes at the expense of head classes~\cite{zhang2021deep}. Secondly, previous research has focused on balancing either classifier learning or representation learning individually, neglecting the importance of simultaneously addressing both aspects. Therefore, it is desirable to jointly optimize representation learning and classifier learning and avoid damaging the head class performance to develop effective and robust long-tailed classification algorithms. 

We identify insufficient training samples and a lack of within-class variability in tail classes as the most significant reason that hinders long-tailed graph classification. Therefore, we propose a novel framework named Retrieval Augmented Hybrid Network (\method{}), which adopts a decoupled training procedure to boost the learning of both the feature extractor and the classifier. In contrast to existing class re-balancing approaches, our framework aims to explicitly introduce additional knowledge to enrich tail classes and learn the balanced weights of the classifier. For the learning of feature extractors, we jointly train a standard base encoder, an additional retrieval branch, and a contrastive learning based branch. Specifically, the retrieval branch makes use of training graphs as retrieval keys and returns the most relevant corpus graphs by subgraph matching techniques, which directly enriches the intra-class diversity for the tail classes. We also explore supervised contrastive learning strategies and tailor them to balance the lower bound of loss value among head and tail classes by introducing a set of category centers \cite{cui2021parametric}. By this means, our balanced supervised contrastive learning (\methodcl{}) aggregates the samples of each category to their center and facilitates better representation learning from long-tailed data. Moreover, we investigate different weight regularizers and use them to fine-tune the classifier, which avoids biased weight norms toward head classes. The contributions of this work can be summarized as follows:

\begin{itemize}[leftmargin=*]
    \item We elaborate on a hybrid network \method{} for long-tailed graph classification, which comprises a retrieval augmentation branch and a balanced contrastive learning module to explicitly introduce intra-class variability and enhance feature learning. 
    \item To balance classifier weights without harming representation learning, we propose to decouple feature extractor learning and classifier training, and fine-tune the classifier with multiple weight regularization techniques.
    \item \method{} is compared with multiple imbalance handling methods on a variety of benchmarks with different imbalance factors. Experimental results show the superiority of our approach in long-tailed graph classification.
\end{itemize}

\section{Related Work}

\subsection{Long-tailed learning} 
Long-tailed learning approaches can be roughly divided into three main categories: class re-balancing, transfer learning, and decoupling training. Generally, the re-balancing strategy aims to equalize the contribution of different classes and emphasize the tail classes from various angles. Specifically, re-sampling techniques ~\cite{zang2021fasa,kim2020imbalanced} balance the classes from the perspective of training data, which achieve a more balanced data distribution across classes. Cost-sensitive learning approaches ~\cite{he2022relieving,tan2021equalization} introduce balance from a different perspective, which adjust loss values for each class during training. However, the use of re-balancing techniques may lead to under-fitting for head classes and reduced accuracy \cite{wang2021margin}. For transfer learning approaches, head-to-tail knowledge transfer seeks to transfer knowledge from data-abundant classes to data-poor classes ~\cite{liu2020deep,park2022majority}, while ~\cite{xiang2020learning,wang2021long} propose to learn a unified student model using adaptive knowledge distillation from the multiple teacher experts. 
Besides, decoupling training methods divide the training procedure into multiple stages. For instance, \cite{kang2020decoupling} proposes to decouple the learning procedure into the encoder learning and the classifier training, and \cite{zhou2020bbn} suggests learning from both the instance-balanced sampling branch and reversed sampling branch.
While some recent work ~\cite{wang2022gog} has applied imbalance handling techniques to imbalanced graph classification, the effectiveness of these methods for multi-class long-tailed graph classification remains uncertain. Our proposed framework simultaneously enhances feature learning with transfer learning and strengthens the classifier training with weight regularization in a decoupled manner to tackle the long-tailed graph classification problem.

\subsection{Graph retrieval} Given a query graph, graph retrieval systems aim to determine the most similar graph among a set of corpus graphs. Recently, leveraging neural architectures to tackle the task of graph matching has received increasing attention. For example, GMN \cite{li2019graph} computes a similarity score for a pair of graphs through a cross-graph attention mechanism to associate nodes across graphs. More recently, ISONet \cite{roy2022interpretable} is also built upon MPNNs to obtain edge embeddings, which are subsequently used to learn an edge alignment network to approximate the underlying correspondence between graphs.

\subsection{Contrastive learning} Contrastive learning has attracted increasing attention in the field of unsupervised representation learning, which aims to aggregate semantically similar samples and obtain discriminative representations by comparing positive and negative pairs. Among them, SimCLR \cite{chen2020simple} and MoCo \cite{he2020momentum} are two classical approaches to self-supervised contrastive learning. Supervised contrastive learning (SCL) \cite{khosla2020supervised} is an extension to contrastive learning by leveraging the label information to compose positive and negative samples, which leads to remarkable performance for classification. 

\section{Methodology}

\begin{figure*}[t]
 \includegraphics[width=1.0\textwidth]{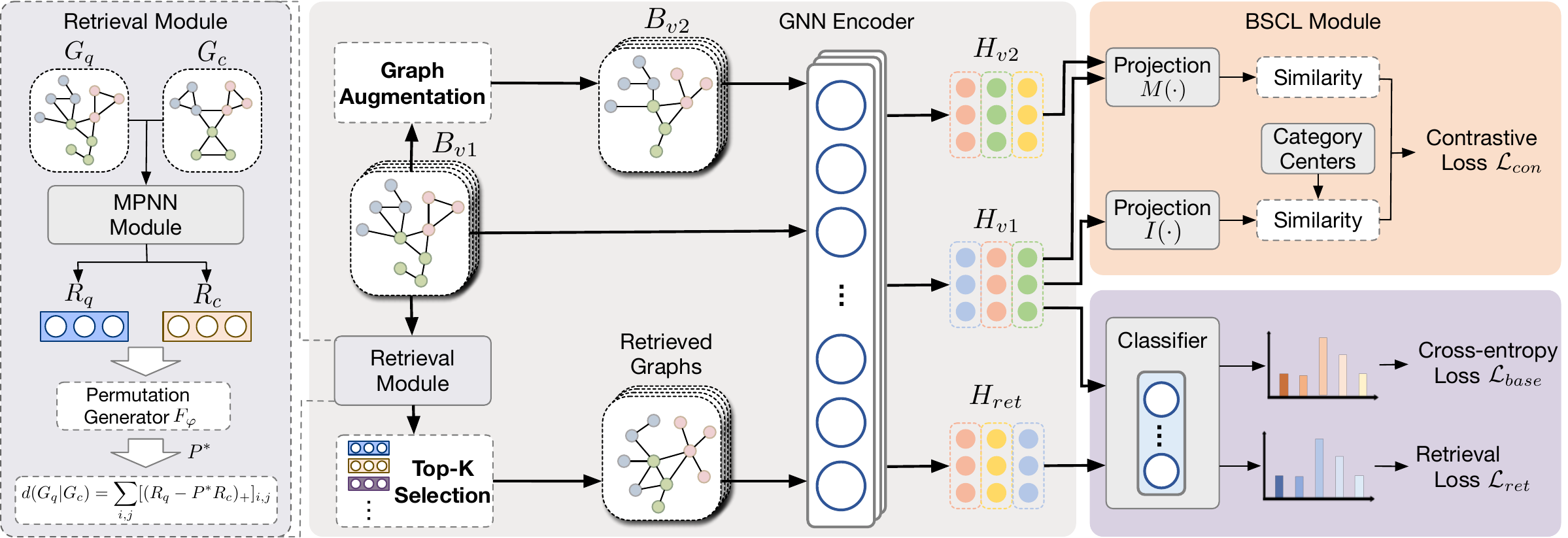}

\caption{Overview of the proposed \method{} architecture. Our \method{} is a hybrid network with three primary branches and corresponding loss values, \ie, the base supervised learning branch ($\mathcal{L}_{base}$), the retrieval augmentation branch ($\mathcal{L}_{ret}$), and the balanced contrastive learning branch ($\mathcal{L}_{con}$), the model is trained by the losses summed up from each branch with different weight coefficients.}
\label{fig:model}

\end{figure*}

This section begins by introducing the problem setting and providing necessary background information. Subsequently, we introduce the proposed \method{} framework (as shown in Figure \ref{fig:model}) in detail, including the retrieval module, \methodcl{} module and classifier regularization. Furthermore, the training pipeline and time complexity analysis of \method{} are thoroughly explained.

\subsection{Preliminaries}
\noindent \textbf{Problem Formulation.} Let $\mathcal{G} = \{ (V_i, E_i, y_i)\}_{i=1}^N$ denote the long-tailed graph dataset, where each graph consists of a node set $V_i$, an edge set $E_i$, along with the corresponding label $y_i \in \{ 1, 2, \ldots, C\}$. The total number of training samples over $C$ classes is $N = \sum_{c=1}^{C} n_{c} $, where $n_{c}$ denotes the data number of class $c$. Without loss of generality, we assume $n_1 \geq n_2 \geq \ldots \geq n_C$, and $n_1 \gg n_C$ after sorting the classes by cardinality in decreasing order, and then the imbalance factor (IF) can be defined as $n_1 / n_C$. In the task of long-tailed graph classification, we aim to learn an unbiased model on a long-tailed training dataset that generalizes well on a balanced test dataset.

\noindent \textbf{Base Encoder.} In our hybrid framework, we adopt a unified GNN encoder shared by multiple branches, which leverages both node attributes and graph topology to obtain embeddings. Let $f(\cdot)$ be a $L$-layer GNN encoder, and $\mathcal{A}^{(l)}_{\theta}$ and $\mathcal{C}^{(l)}_{\theta}$ denote the aggregation and combination functions at the $l$-th layer, then the propagation of the $l$-th layer is formulated as:
\begin{equation} 
    \mathbf{h}_{v}^{(l)}= \mathcal{C}^{(l)}_{\theta}\left(\mathbf{h}_{v}^{(l-1)}, \mathcal{A}^{(l)}_{\theta} \left(\left\{\mathbf{h}_{u}^{(l-1)}\right\}_{u \in \mathcal{N}(v)}\right) \right),
\end{equation}
where $\mathbf{h}_{v}^{(l)}$ is the embedding of the node $v$ at the $l$-th layer, $\mathcal{N}(v)$ refers to the neighbors of $v$. After $l$ iterations, the final graph embedding $f(G)$ is integrated from embedding vectors at all layers through a $\textit{READOUT}$ function:
\begin{equation}
   \mathbf{h}_i = f(G_i) = \textit{READOUT}(\{\mathbf{h}_{v}^{(l)}: v \in \mathcal{V}_i, l \in L\}).
\end{equation}

\subsection{Retrieval Augmented Branch}
In the task of graph retrieval, we are given a set of query graphs $G_q$ and a set of corpus graphs $G_c$, the goal is to design a neural distance function $d(G_q | G_c)$ that accurately predicts the similarity between them. We first pad $G_q$ with $\left| V_c \right| - \left| V_q \right|$ dummy nodes, and the augmented node adjacency matrices of size $W$ can be represented as $\mathbf{A_q}$ and $\mathbf{A_c}$. The objective function for the node alignment approach is defined as:
\begin{equation}
\label{eq:nodeAlign}
    \mathop{\arg\min}\limits_{S} \sum\limits_{i,j \in \left[ W \right] \times \left[ W \right]} \left[ \big (\mathbf{A_q} - \mathbf{S} \mathbf{A_c} \mathbf{S^{\top}} \big )_{+} \right]_{i,j},
\end{equation}
where $(\cdot)_{+} = \max \{0,\cdot \}$, $\left[ \cdot \right]_{i,j}$ represents the element located at the $i$-th row and $j$-th column of the matrix, and $\mathbf{S}$ is the node permutation matrix. The node alignment methods seek to approximate Eq. (\ref{eq:nodeAlign}) with node embeddings, which is still a hard quadratic assignment problem. Thus, one feasible way of node alignment \cite{li2019graph} is directly aggregating node embeddings to graph embedding and performing graph matching, which may fail to capture edge correspondences. To tackle this problem, we perform subgraph matching with edge alignment by rewriting the objective of Eq.~\ref{eq:nodeAlign} as:
\begin{equation}
\label{eq:edgeAlign}
    \mathop{\arg\min}\limits_{S} \sum\limits_{k \in \left[ W^2 \right]} \left[ \big ( {\rm vec} \left( \mathbf{A_q} \right) - \left( \mathbf{S} \otimes \mathbf{S} \right) {\rm vec} \left( \mathbf{A_c} \right) \big )_{+} \right]_{k},
\end{equation}
where $\otimes$ is the Kronecker product and ${\rm vec} (\mathbf{A_{\cdot}})$ denotes the vector obtained by column-wise concatenation. In practice, we approximate Eq.~\ref{eq:edgeAlign} with edge embeddings. Let $\mathbf{r_e} \in \mathbb{R}^D$ denote the embedding of edge $e$, and the edge embedding matrices can be represented as $\mathbf{R_q}, \mathbf{R_c} \in \mathbb{R}^{\left| E \right| \times D}$ collectively. Therefore, we can write the distance measure as:
\begin{equation}
\label{eq:disEdge}
    d(G_q | G_c) = \min_{P} \sum_{e} \left[ \big (\mathbf{R_q} - \mathbf{P} \mathbf{R_c} \big )_{+}\right]_e,
\end{equation}
where $\mathbf{P}$ is an edge permutation without a quadratic term.

We develop our edge alignment-based retrieval branch by approximate $d(G_q | G_c)$ in Eq.~\ref{eq:disEdge} with two neural networks as $d_{\phi, \varphi}(G_q | G_c)$: (i) The first network is a message passing framework with parameter $\phi$ to model the edge embedding matrices $\mathbf{R_q}$ and $\mathbf{R_c}$, (ii) The second network with parameter $\varphi$ is the Gumbel-Sinkhorn operator, which aims to provide a differentiable solution $\mathbf{P^*}$ to address the optimization problem. Therefore, the search for discrete edge permutation matrix $P$ is replaced with the relaxation:
\begin{equation}
    d_{\phi, \varphi}(G_q | G_c) = \sum_{i,j} \left[ \big (\mathbf{R_q} - F_{\varphi} \left( \mathbf{R_q}, \mathbf{R_c} \right) \mathbf{R_c} \big )_{+}\right]_{i,j}.
\end{equation}

Finally, we follow \cite{roy2022interpretable} to train the above two networks with parameters of $\phi$ and $\varphi$ through a hinge loss:
\begin{equation}
\resizebox{.91\linewidth}{!}{$
    \min_{\phi, \varphi} \sum\limits_{G_q} \sum\limits_{G_{c\oplus},G_{c\ominus}} \left[ \gamma + d_{\phi, \varphi}(G_q | G_{c\oplus}) - d_{\phi, \varphi}(G_q | G_{c\ominus}) \right]_{+},
    $}
\end{equation}
where $\gamma \in \mathbb{R}^{+}$ is a margin parameter, and $G_{c\oplus}$ and $ G_{c\ominus}$ represent the relevant and irrelevant graphs for $G_q$, respectively. We use a breadth-first search-based technique to sample $G_{c\oplus}$ and $ G_{c\ominus}$ from $\mathcal{G}$, employing the VF2 exact match algorithm \cite{lou2020neural} to determine the ground-truth training targets for any given $G_q$.

With the pre-trained retrieval network $d_{\phi, \varphi}(\cdot | \cdot)$, we set both $G_q$ and $G_c$ to be the training dataset, and retrieve each query graph for top-$q$ relevant graphs. In the implementation, we augment the original data batch $B$ to $\hat{B}$ with retrieved graphs, $\hat{B}$ is then fed into the encoder to obtain both the base representation $\mathbf{h}_{base}$ and representations of retrieved graphs $\hat{\mathbf{h}} \in \mathbb{R}^{q \times D}$, where $D$ is the dimension of embedding vectors. The retrieval feature $\mathbf{h}_{ret}$ are then adaptively combined by the representations of $q$ retrieved graphs, each with a different weight coefficient:
\begin{equation}
    \mathbf{h}_{ret} = \mathbf{a}^{\top} \hat{\mathbf{h}} = {\sum_{i = 1}^{q}{\mathbf{a}_{i} \hat{\mathbf{h}}_{i}}},
\end{equation}
where $\mathbf{a} \in \mathbb{R}^{q}$ represents the attention coefficients over each retrieved graph derived from the base feature. Specifically, we use a fully-connected layer $\mathbf{W}_{a}$ with a softmax activation function to obtain the attention coefficients: $\mathbf{a} = Softmax\left( \mathbf{W}_{a} \mathbf{h}_{base} \right) $. The obtained embeddings $\mathbf{h}_{base}$ and $\mathbf{h}_{ret}$ are forwarded by a classifier layer to obtain the prediction probabilities. Finally, the prediction probabilities for both the supervised learning branch and retrieval augmented branch are used to calculate the standard cross-entropy loss, which can be denoted as $\mathcal{L}_{base}$ and $\mathcal{L}_{ret}$, respectively.

\subsection{Balanced Supervised Contrastive Learning}
In this subsection, we discuss how to leverage contrastive learning to produce discriminative and robust representations under the long-tailed setting. Following ~\cite{you2020graph,cui2021parametric}, we first obtain different positive views of graphs by involving four fundamental data augmentation strategies that preserve intrinsic structural and attribute information: \textit{(1) Edge permutation (2) Attribute masking (3) Node dropping (4) Subgraph}. During training, we are given a two-viewed graph batch $B=(B_{v1}, B_{v2})$ and label $y$, the graphs in $B_{v2}$ are transformed by aforementioned data augmentation strategies. Each view of graphs is fed into the shared GNN encoder to obtain the embeddings $\mathbf{H_{v1}}$ and $\mathbf{H_{v2}}$, respectively. Afterward, the embeddings are fed into the projection network $g(\cdot)$ which projects $H$ to another latent space where the contrastive loss is calculated. After obtaining projected representation $\mathbf{Z_{v1}}$ and $\mathbf{Z_{v2}}$, we define the positives $P(i)$ for any anchor graph $G_i$ in $B_{v1}$ as:
\begin{equation}
    \begin{aligned}
        A(i) &= \{\mathbf{z_k} \in \mathbf{Z_{v1}} \cup \mathbf{Z_{v2}}\} \backslash \{\mathbf{z_k} \in \mathbf{Z_{v1}} : k=i\}, \\
        P(i) &= \{\mathbf{z_k} \in A(i): y_k = y_i\}.
    \end{aligned}
\end{equation}

However, the scarcity of tail classes will result in the lack of positive pairs, which consequently leads to performance deterioration of instance-level contrastive loss on the long-tailed dataset. Therefore, we incorporate a set of category centers $\mathbf{O}$ to balance the loss value among head and tail classes. More specifically, the category centers are a set of learnable parameters, and the dimension of $\mathbf{O}$ is $\mathbb{R}^{C \times D}$, where $C$ is the number of classes and $D$ is the embedding dimension. Formally, the loss of \methodcl{} can be written as:
\begin{equation}
\label{eq:cl}
    \resizebox{.91\linewidth}{!}{$
        \displaystyle
        \mathcal{L}_i^{con} = \sum\limits_{\mathbf{z_{+}} \in P(i) \cup {\mathbf{o}_y}} -w(z_{+}) \log \frac{exp(\mathbf{z_{+}} \cdot g(\mathbf{h_i}) / \tau)} {\sum_{\mathbf{z_k} \in A(i) \cup \mathbf{O}} exp(\mathbf{z_k} \cdot g(\mathbf{h_i}) / \tau)},
    $}
\end{equation}
where $\tau \in \mathbb{R}^{+}$ is a scalar temperature parameter, and
\begin{equation}
    w(\mathbf{z_{+}}) = \left\{
    \begin{aligned}
         1.0,\quad&\mathbf{z_{+}} \in {\mathbf{o}_y} \\
         \alpha,\quad&\mathbf{z_{+}} \in P(i)
    \end{aligned}
    \right..
\end{equation}

Moreover, we instantiate $g(\cdot)$ as either a multi-layer perceptron ${M}(\cdot)$ or an identity mapping layer ${I}(\cdot)$, \ie, ${I}(x) = x$. Specifically, different projection head is chosen as follows:
\begin{equation}
    \mathbf{z} \cdot g(\mathbf{\mathbf{h_i}}) = \left\{
    \begin{aligned}
         &\mathbf{z} \cdot {I}(\mathbf{h}_i), \quad \mathbf{z} \in \mathbf{O} \\
         &\mathbf{z} \cdot {M}(\mathbf{h}_i), \quad \mathbf{z} \in A(i)
    \end{aligned}
    \right..
\end{equation}

To demonstrate the superiority of \methodcl{} in long-tailed learning, we further conduct an analysis of the loss value between supervised contrastive learning and \methodcl{}. Suppose $K_{y_i}$ is the expected number of positive pairs with respect to given graph $G_i$ and its label $y_i$, which can be calculated as:
\begin{equation}
    K_{y_i} = (2 * \textit{batchsize} - 1) * \pi_{y_i},
\end{equation}
where $\pi_{y_i}$ is the class frequency over the whole dataset, \ie, $n_{y_i}/N$. When supervised contrastive loss achieves minimum, the optimal value for the probability that two graph samples are a true positive pair is $1/K_y$, where $y$ is the corresponding label. In that case, the head classes will have a higher lower bound of loss value and contribute significantly more importance than tail classes during training. However, in regard to \methodcl{} loss, the optimal value for the probability that two graph samples are a true positive pair is $\alpha / ({1 + \alpha \cdot K_y})$, and optimal value for the probability that a graph is closest to its category center $\mathbf{o}_y$ is $1 / ({1 + \alpha \cdot K_y})$. Therefore, the head-to-tail optimal value gap is reduced from $1 / {K_{y_{head}}} \rightarrow 1 / {K_{y_{tail}}}$ to $1 / ({1 / \alpha + K_{y_{head}}}) \rightarrow 1 / ({1 / \alpha + K_{y_{tail}}})$. As the value of $\alpha$ decreases, the optimal value from head to tail becomes increasingly balanced, which enhances the representation quality of \methodcl{} in the long-tailed settings.

\begin{algorithm}[t]
    \caption{Training algorithm of \method{}}
    \label{alg:training}
    \textbf{Input}: Training set $\mathcal{G}$ \\
    \textbf{Parameter}: GNN encoder $f(\cdot,\theta_f)$, pre-trained retrieval network $d_{\phi, \varphi}(\cdot | \cdot)$, contrastive learning module parameter $\psi (\cdot, \theta_b)$, classifier $(\cdot, \theta_c)$, epoch number $T$, epoch number for fine-tuning $T_f$. \ \\
    \textbf{Output}: The proposed \method{}
    \begin{algorithmic}[1] %[1] enables line numbers
        \FOR{$t = 1$ to $T$}
        \STATE Sample batch $B$ from the instance-balanced sampler.
        \STATE Augment $B$ to a two-viewed batch $B=(B_{v1},B_{v2})$.
        \STATE Retrieve for relevant graphs $G_r$ using $d_{\phi, \varphi}(\cdot | \cdot)$.
        \STATE Expand $B_{v1}$ to $\hat{B}_{v1}$ with $G_r$.
        \STATE Forward propagation $\hat{B}_{v1}$ via $f(\cdot,\theta_f)$ and $(\cdot, \theta_c)$.
        \STATE Compute $\mathcal{L}_{base}$ and $\mathcal{L}_{ret}$.
        \STATE Forward propagation $B=(B_{v1},B_{v2})$ via $\psi (\cdot, \theta_b)$.
        \STATE Compute contrastive loss $\mathcal{L}_{con}$ using Eq.~\ref{eq:cl}.
        \STATE Sum up the loss values using Eq.~\ref{eq:total}.
        \STATE Update $\theta_f$, $\theta_b$, and $\theta_c$ by back propagation.
        \ENDFOR
        \STATE Fix the weights of GNN encoder $\theta_f$.
        \FOR{$t = 1$ to $T_f$}
        \STATE Sample batch $B$ using the class-balanced sampler.
        \STATE Forward propagation $B$ via $f(\cdot,\theta_f)$ and $(\cdot, \theta_c)$.
        \STATE Compute the objective function of cross-entropy and weight decay.
        \STATE Update $\theta_c$ by back propagation.
        \STATE Balance $\theta_c$ using equation Eq.~\ref{eq:pgd}.
        \ENDFOR
        % \STATE \textbf{return} \method{}
    \end{algorithmic}
\end{algorithm}

\subsection{Re-balancing the Classifier}
Long-tailed class distribution often results in larger classifier weight norms for head classes \cite{yin2019feature}, which makes the classifier easily biased to dominant head classes. To address this problem, we regularize the weights of the classifier with the trained feature extractor frozen. Specifically, we investigate two different weight regularization approaches to balance weights with respect to norms, \ie, Max-norm and weight decay.

\paragraph{Max-norm.} Max-norm constrains the weights to have a norm less than or equal to a specific value. More formally, Max-norm caps weight norms within an L2-norm ball:
\begin{equation}
\label{eq:maxnorm}
    \Theta^{*} = \mathop{\arg\min}\limits_{\Theta} F(\Theta; \mathcal{G}), \quad \mathrm{ s.t. } \quad \Vert \theta_{k} \Vert_2^2 \leq \delta^2, \ \forall k,
\end{equation}
where $\delta$ is the radius. Moreover, we adopt projected gradient descent (PGD) to solve Eq.~\ref{eq:maxnorm}, which efficiently projects out-of-ranged weights onto the L$2$-normball. Concretely, it applies a renormalization step at each iteration. After each batch update, PGD computes an updated $\theta_k$ and projects it onto the L2-norm ball with radius $\delta$:
\begin{equation}
\label{eq:pgd}
    \theta_k \leftarrow \min(1, \delta / \Vert \theta_{k} \Vert_2) * \theta_{k}.
\end{equation}

Unlike L2-normalization, which forces weights to be unit-length in norm, Max-norm loosens this constraint, allowing the weights to fluctuate within the norm-ball during training.

\paragraph{Weight Decay.} Weight decay is a widely used type of regularization, which is utilized to constrain the growth of the network weights. Weight decay typically penalizes the network weights according to their L2-norm:
\begin{equation}
    \Theta^{*} = \mathop{\arg\min}\limits_{\Theta} F(\Theta; \mathcal{G}) + \lambda \sum \limits_k \Vert \theta_{k} \Vert_2^2,
\end{equation}
where the hyperparameter $\lambda$ is used to control the impact of weight decay. It enhances the model's generalization performance by punishing large weights and encouraging learning small balanced weights which consequently decreases the complexity of the network to prevent overfitting.

Moreover, jointly applying Max-norm and weight decay leads to better performance because of their complementing benefits. Generally, Max-norm caps large weights within the specific radius and prevents them from dominating the training, while weight decay on the small weights still improves the overall generalization and avoids overfitting. 

\begin{table}[t]
\centering
\caption{Statistics of the datasets used in the experiments.}
\label{table:datasets}
\resizebox{0.95\columnwidth}{!}{
\begin{tabular}{ccccc}
    \toprule 
    \midrule
    Dataset & \# Graphs & \# Classes & \# Avg. Nodes & \# Avg. Edges \\
    \midrule
    Synthie &400 &4 &95.00	&172.93 \\
    ENZYMES &600 &6	&32.63	&62.14 \\
    MNIST & 60,000 & 10 & 75 & 1393.27  \\
    Letter-High & 2250 & 15 &4.67 &4.50  \\
    Letter-low & 2250 & 15 &4.67 &4.50  \\
    COIL-DEL & 3900 & 100 & 21.54 &54.24  \\
    \midrule
    \bottomrule
\end{tabular}
}
\end{table}

\subsection{Training Pipeline}
We train our \method{} framework in a two-stage manner, which decouples representation learning from classifier learning. We also utilized different sampling strategies for different training stages: (i) instance-balanced sampler, in which each data sample has the same probability of being sampled; (ii) class-balanced sampler, in which each class is sampled uniformly and each instance is sampled uniformly within it.

In the first stage, we use an instance-balanced sampler to learn a better feature extractor that preserves better generalizability. Given a data batch $B$, we first transform $B$ to a two-viewed graph batch $B=(B_{v1},B_{v2})$ through data augmentation. The original view of batch $B_{v1}$ is fed into the base branch and retrieval augmented branch to deduce the standard cross-entropy loss $\mathcal{L}_{base}$ and the retrieval loss $\mathcal{L}_{ret}$, respectively. While $B=(B_{v1},B_{v2})$ is fed into the \methodcl{} module to deduce the contrastive loss $\mathcal{L}_{con}$. Finally, \method{} sums up those losses by:
\begin{equation}
\label{eq:total}
    \mathcal{L}_{total} = \mathcal{L}_{base} + \eta_{ret} \cdot \mathcal{L}_{ret} + \eta_{con} \cdot \mathcal{L}_{con},
\end{equation}
where $\eta_{ret}$ and $\eta_{con}$ are hyperparameters that control the contribution of different branches.

In the second stage, a class-balanced sampler together with Max-norm and weight decay regularization are used to learn a balanced classifier. To avoid representation damage because of the balanced sampler, we fix the feature encoder and only fine-tune the classifier layers with cross-entropy loss. The overall training pipeline is illustrated in Algorithm \ref{alg:training}.

\subsection{Time Complexity Analysis}
Denote the batch size $B$, and the average number of nodes in the input graphs is $\left| V \right|$. The time complexity of acquiring embeddings from the GNN encoder for a batch is $O(B L D \left| V \right|)$, where $L$ is the number of layers of the encoder and $D$ denotes the embedding dimension. With the pre-trained retrieval model, the complexity of obtaining the retrieval feature is $O(qB)$, where $q$ is the number of retrievals. For BSCL, we calculate the loss for a batch in $O(DB^2)$ time, and for the classifier re-balancing, the time complexity of max-norm and weight decay is $O(BC)$, where $C$ is the number of classes. Overall, the total time complexity for \method{} is $O\left( B \left( L D \left| V \right| + q + DB + C \right) \right)$.

\section{Experiment}

\begin{table*}[t]
\begin{center}
\caption{Long-tailed graph classification accuracy on six benchmark datasets with various IFs (best performance in bold).}
\label{table:experiment}
\resizebox{0.95\textwidth}{!}{ %
\begin{tabular}{lcccccccccccc}
\toprule
\midrule
Model &\multicolumn{2}{c}{Synthie} &\multicolumn{2}{c}{ENZYMES}  &\multicolumn{2}{c}{MNIST} &\multicolumn{2}{c}{Letter-high} &\multicolumn{2}{c}{Letter-low} & \multicolumn{2}{c}{COIL-DEL}\\
\cmidrule{2-3} \cmidrule{4-5} \cmidrule{6-7} \cmidrule{8-9} \cmidrule{10-11} \cmidrule{11-13}
& IF=15 & IF=30 & IF=15 & IF=30  & IF=50 & IF=100 & IF=25 & IF=50  & IF=25 & IF=50 & IF=10 & IF=20\\ 
\midrule
GraphSAGE       &34.74  &30.25  &30.66  &25.16  &68.67  &63.46  &51.06  &42.16  &86.00  &84.32  &38.80  &31.32 \\
Up-sampling      &35.25  &33.50  &32.33  &28.50  &64.69  &59.78  &53.62  &44.20  &88.48  &86.72  &39.20  &26.96 \\
\midrule
CB loss         &34.75  &30.75  &32.19  &26.83  &68.85  &63.40  &53.76  &45.06  &87.46  &85.44  &41.72  &32.34 \\
LACE loss       &33.25  &30.85  &31.16  &25.50  &69.72  &64.59  &47.46  &38.94  &87.89  &84.69  &41.96  &32.18 \\
\midrule
Augmentation    &39.37  &35.37  &32.08  &26.75  &72.18  &68.17  &49.28  &42.36  &88.32  &86.40  &38.18  &30.80 \\
G$^2$GNN$_{n}$     &38.08  &27.94  &35.00  &29.17  &70.91  &66.73  &58.91  &51.12  &89.49  &87.98  &38.32  &27.98 \\
G$^2$GNN$_{e}$     &40.19  &37.53  &35.83  &29.50  &73.69  &70.31  &58.85  &49.96  &89.84  &87.80  &39.18  &31.06 \\ 
\midrule
GraphCL     &40.25   &36.25   &36.66  &29.83    &69.37  &65.12  &57.34  &48.93  &89.28  &87.89  &42.02   &33.19 \\
SupCon      &40.34   &37.25   &37.08  &30.67    &69.76  &64.88  &57.29  &48.93  &89.12  &87.36  &42.93   &34.20 \\
\midrule
\method{}       &\textbf{42.35}  &36.76  &38.50  &32.17  &75.12  &71.98  &59.20  &50.37  &89.65  &88.69  &43.04  &36.80 \\
\method{}$_{dec}$ &40.75 &\textbf{39.00}  &\textbf{39.02}  &\textbf{34.16}  &\textbf{75.79}  &\textbf{72.60}  &\textbf{59.79}  &\textbf{52.90}  &\textbf{90.19}  &\textbf{89.28}  &\textbf{45.32}  &\textbf{38.48} \\
\midrule
\bottomrule
\end{tabular}
}

\end{center}
\end{table*}

\subsection{Experimental Setups}

\smallskip\noindent\textbf{Benchmark Datasets.} We evaluate our proposed \method{} on six publicly accessible datasets in various fields, including synthetic (Synthie \cite{morris2016faster}), bioinformatics (ENZYMES \cite{schomburg2004brenda}), and computer vision (MNIST \cite{dwivedi2020benchmarking}, Letter-high \cite{riesen2008iam}, Letter-low \cite{riesen2008iam}, and COIL-DEL \cite{riesen2008iam}). We follow Zipf’s law to process the original datasets into long-tailed datasets. Moreover, we split the dataset into train/val/test sets in a ratio of 60\%/20\%/20\%, respectively.

\smallskip\noindent\textbf{Competing Models.} We carry out comprehensive comparisons with methods from four categories, which include: (a) data re-balancing method, \ie, up-sampling \cite{chawla2003c4}, (b) loss re-weighting approaches, \ie, class-balanced (CB) loss \cite{cui2019class}, and logit adjustment cross-entropy (LACE) loss \cite{menon2020long}, (c) information augmentation, \ie, data augmentation \cite{yu2022graph}, and G$^2$GNN \cite{wang2022gog}, and (d) contrastive learning, \ie, GraphCL \cite{you2020graph}, and SupCon \cite{khosla2020supervised}.

\smallskip\noindent\textbf{Implementation Details.} 
For the proposed \method{},  we use GraphSAGE~\cite{ahmed2017inductive} as the base GNN encoder, and empirically set the embedding dimension to $64$, the number of epochs to $1000$, and batch size to $32$. The softmax temperature $\gamma$ is set to $0.2$, the contrast weight $\alpha$ is set to $0.05$, and weight decay of the second stage is $0.1$. 
Moreover, we tune $\eta_{ret}$, $\eta_{con}$ and $\delta$ for different datasets, and average accuracy over $10$ runs is used as the evaluation metric.

\subsection{Results and Analysis}
The experimental results on the six benchmarks with different imbalance factors (IFs) are shown in Table \ref{table:experiment}. Our experiment involves comparing the baseline methods with two variants of our \method{} model: one without weight regularization (\method{}) and the other with decoupled classifier training (\method{}$_{dec}$). We can make the following observations: 
\begin{itemize}[leftmargin=*]
    \item The classification performance of both baseline methods and our proposed \method{} suffers a sharp decrease in all six datasets when the long-tailedness between head and tail classes increases, indicating that GNNs often degrade and result in sub-optimal classification performance under the long-tailed setting.
    \item Among the four categories of baselines, information augmentation approaches surpass data re-sampling and loss re-weighting baselines on most datasets. This may be due to the fact that it is crucial for information augmentation methods to explicitly introduce new knowledge to enrich the tail classes, leading to better representation ability for the tail. Moreover, contrastive learning-based approaches exhibit relatively stable performance across all six datasets. 
    \item Our \method{} and its variant (\method{}$_{dec}$) consistently outperform other baselines on all six datasets under different imbalance settings, especially under harsh imbalance settings such as COIL-DEL dataset with IF=$20$, where over $60$ classes only contain less than $3$ training samples. Moreover, even without re-balancing the classifier, our \method{} still outperforms the others in most scenarios, which implies that the retrieval branch and \methodcl{} module jointly learn discriminative representations and play a significant role in alleviating the long-tailedness problem.
\end{itemize}

\begin{table}[t]
\begin{center}
\caption{Ablation study for the primary components of \method{} (RA: Retrieval Augmentation, SCL: Supervised Contrastive Learning, BSCL: Balanced Supervised Contrastive Learning, and WR: Weight Regularization).}
\label{table:ablation}
\resizebox{0.95\columnwidth}{!}{
\begin{tabular}{cccc|cc}
\toprule
\midrule
RA     &SCL     &\methodcl{}    &WR     &Letter-high    &ENZYMES \\
\midrule
\Checkmark  &\ & & &48.83 &36.67\\
  &\Checkmark & & &48.93 &37.08\\
  & &\Checkmark & &49.51 &37.78\\
\Checkmark &\Checkmark & & &49.58 &37.83\\
 &\Checkmark & &\Checkmark &50.76 &37.61\\
\Checkmark  & &\Checkmark & &50.37 &38.50\\
\Checkmark  & &\Checkmark &\Checkmark &52.90 &39.02\\
\midrule
\bottomrule
\end{tabular}}

\end{center}
\end{table}

\begin{figure}[t]
 \includegraphics[width=0.48\textwidth]{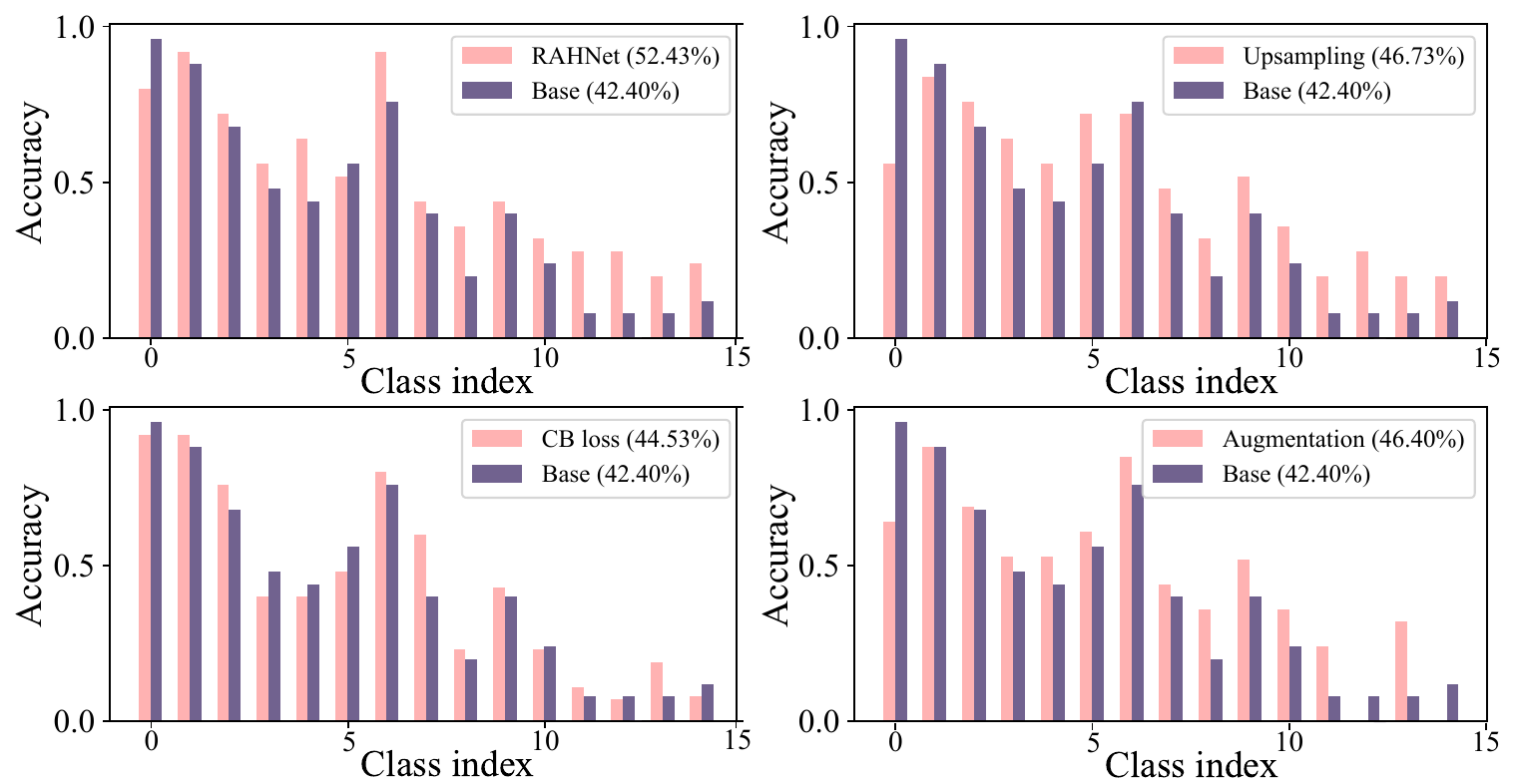}

\caption{Per-class classification accuracy on the Letter-high dataset with an IF of 50.}
\label{fig:ablation}
%
% \vspace{-0.3cm}
\end{figure}

\begin{figure}[t]\small
\centering
\subfigure[Max-norm threshold (Letter-high)]{
\label{fig:param_a}
\centering
\includegraphics[width=0.22\textwidth]{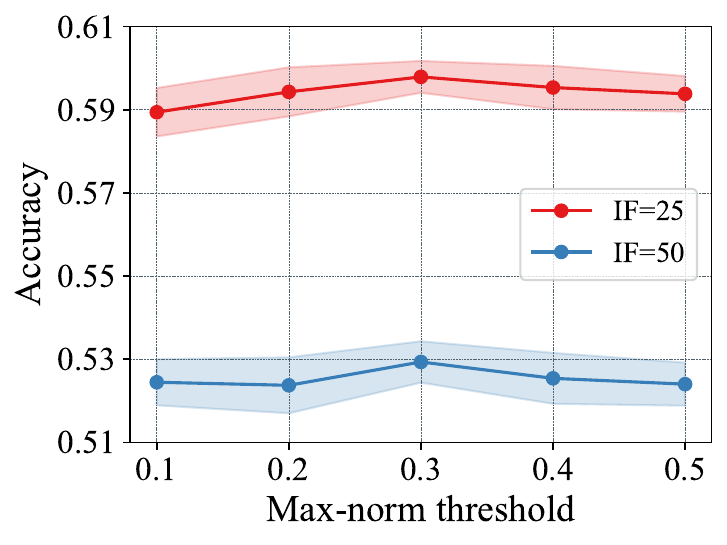}
}
% \hspace{-2mm}
\subfigure[Max-norm threshold (COIL-DEL)]{
\label{fig:param_b}
\centering
\includegraphics[width=0.22\textwidth]{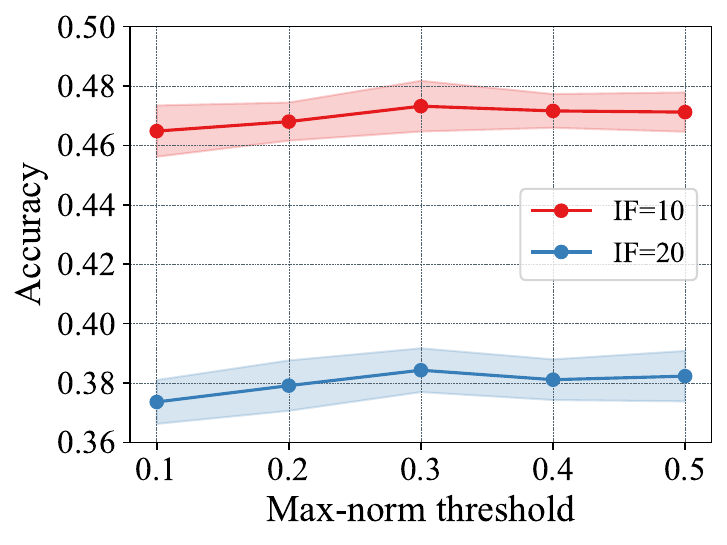}
}

\subfigure[Loss weight (Letter-high)]{
\label{fig:param_c}
\centering
\includegraphics[width=0.22\textwidth]{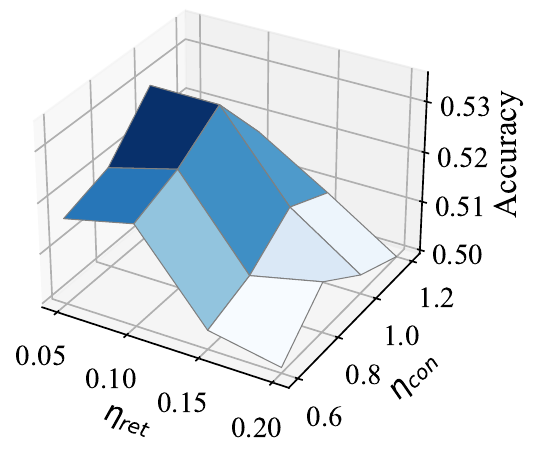}
}
% \hspace{-2mm}
\subfigure[Loss weight (COIL-DEL)]{
\label{fig:param_d}
\centering
\includegraphics[width=0.22\textwidth]{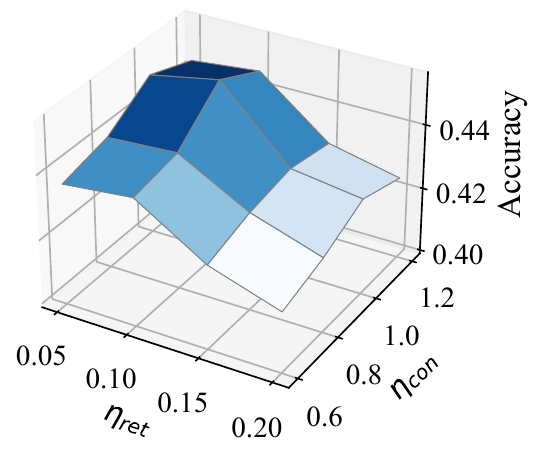}
}

\caption{Hyper-parameter sensitivity study of \method{}.}
\label{fig::parameter}
\end{figure} 

\subsection{Ablation Study}

We perform several ablation studies to characterize the proposed \method{}. First, we compare the performance of the primary components of the framework in Table \ref{table:ablation}, where retrieval augmentation (RA), balanced supervised contrastive learning (\methodcl{}), and weight regularization (WR) are three main components in \method{} framework. Moreover, we investigate the effectiveness of \methodcl{} by substituting the supervised contrastive learning (SCL) module for \methodcl{}. All experiments are performed on Letter-high (IF=$50$) and ENZYMES (IF=$15$). First of all, we can clearly observe that using both retrieval augmentation and balanced contrastive learning solely can enhance the overall performance, and jointly training the model with both of the components can significantly boost the performance. Secondly, combining \methodcl{} with other components (RA + \methodcl{}) outperforms RA + SCL. The results indicate that SCL is not applicable to long-tailed datasets due to the lack of positive pairs for tail classes, and our proposed \methodcl{} effectively balanced the training in long-tailed distributions. Finally, our complete framework achieves better performance, demonstrating that weight regularization helps to learn an unbiased classifier under the long-tailed setting.

We further study the impact of decoupling training by comparing the per-class accuracy, the results are shown in Figure \ref{fig:ablation}. It can be observed that all the approaches, perform well on data-rich head classes but significantly worse on the tail classes. Our \method{} framework enhances both head and tail classes classification accuracy with little loss of head-class performance. Without explicitly re-balancing the data samples or loss contribution during training, decoupling training with weight regularization still balances the performance between head and tail classes well.

\subsection{Hyper-parameter Study}

We further study the sensitivity of hyper-parameters in \method{}. First, we examine the effect of Max-norm threshold $\delta$ by varying $\delta$ in \{$0.1$, $0.2$, $0.3$, $0.4$, $0.5$\} with all other hyper-parameters fixed. The experimental results on the Letter-high and COIL-DEL datasets are shown in Figure \ref{fig:param_a} and \ref{fig:param_b}, respectively. We can clearly observe that the classification accuracy rises as the Max-norm threshold increases from $0.1$ to $0.3$, and the performance saturates as the threshold reaches $0.3$ on both datasets with various IFs, which validates that larger $\delta$ offers more free space within the norm ball to let weights grow and also caps the weights to prevent the head class from dominating the training.

Moreover, we study the effect of loss weight hyper-parameters of the retrieval loss $\eta_{ret}$ and the contrastive loss $\eta_{con}$, which are critical to the learning of the feature extractor. We conduct experiments on the Letter-high (IF=50) and COIL-DEL (IF=10) datasets by varying $\eta_{ret} \in \{ 0.05, 0.1, 0.15, 0.2 \}$ and $\eta_{con} \in \{ 0.6, 0.8, 1.0, 1.2\}$. In Figure \ref{fig:param_c} and \ref{fig:param_d}, It can be found that the accuracy reaches the peak when $\eta_{ret}$ is set to $0.1$ and $\eta_{con}$ is set to $1.0$. As $\eta_{ret}$ rises from $0.05$ to $0.1$, the performance improves accordingly.
These results indicate that increasing the retrieval branch contribution of loss in an appropriate range can be beneficial to long-tailed learning because tail classes are augmented with diverse retrieval features.
The \methodcl{} module serves the purpose of representation enhancement and highlights the tail classes during training. Therefore, the effectiveness of \methodcl{} is maximized when the contrastive loss shares an equal contribution of the cross-entropy loss. 

\begin{figure}[t]
 \includegraphics[width=0.48\textwidth]{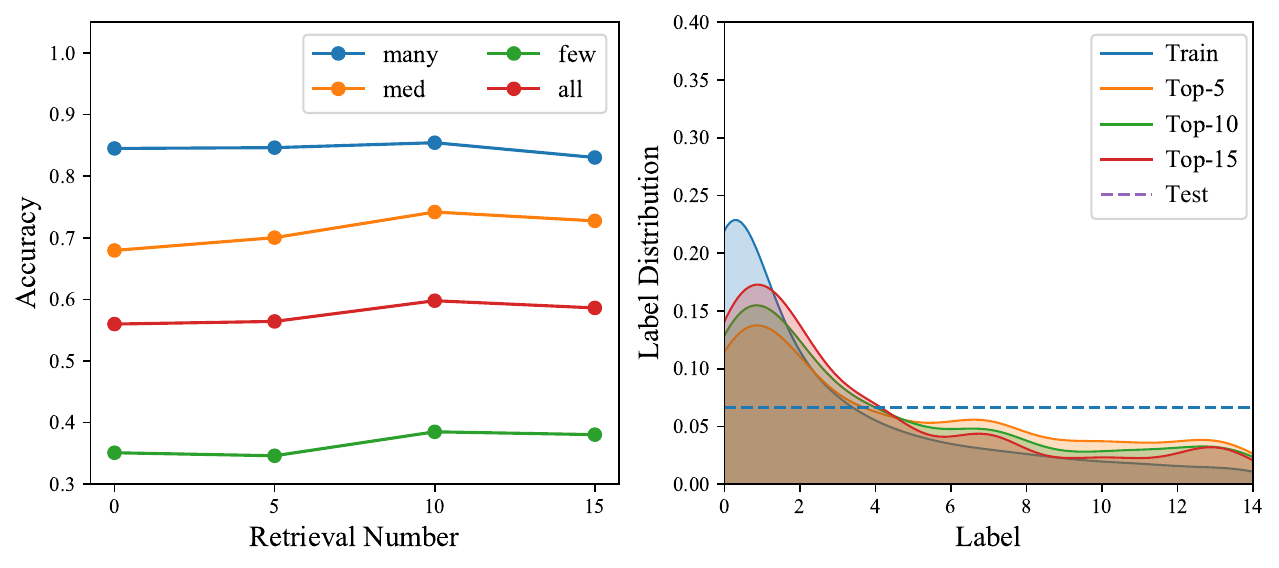}

\caption{Analysis of retrieval module on Letter-high (IF=25). Left: Comparison of using different retrieval number $q$; Right: Label distributions of retrieval augmented training data.
}
\label{fig:ret}
%
% \vspace{-0.3cm}
\end{figure}

\subsection{Retrieval Branch Analysis}
As shown in Figure \ref{fig:ret}, we analyze the effectiveness of the retrieval branch by varying the retrieval number $q$ from $0$ to $15$. We first divide the different classes into many-shot, med-shot, and few-shot categories according to their number of samples, and record the corresponding accuracy when $q$ changes. It can be concluded that the performance on med-shot and few-shot categories consistently increases when $q$ increases from $0$ to $10$, indicating the retrieval module is able to emphasize tail classes during training. Long-tailed learning can also be formulated as a label shift problem where the training and testing label distributions are different. Therefore, we visualize the label distribution of original training data, testing data, and retrieval augmented data. As can be seen from the figure, retrieval augmentation alleviates the class long-tailedness by diminishing the dominance of the head class and making the label distribution closely aligned with the distribution of the test data. The top-10 distribution among various levels of augmentation exhibited the lowest oscillation amplitude in the tail and demonstrated the best overall performance.

\begin{figure}[t]
 \includegraphics[width=0.49\textwidth]{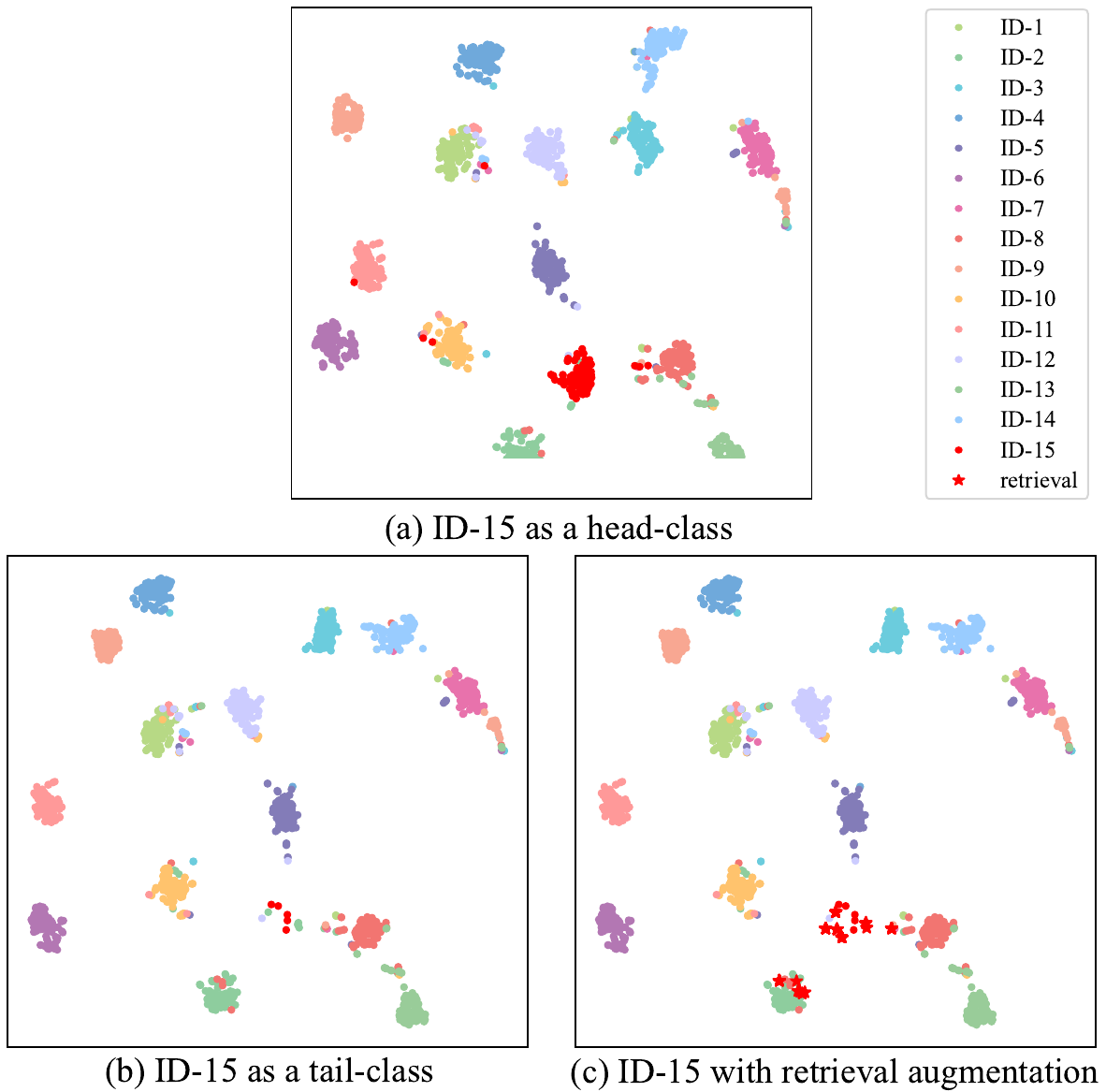}

\caption{Visualize the effectiveness of retrieval augmentation on the COIL-DEL dataset with t-SNE.}
\label{fig:case}

\end{figure}

\subsection{Case Study}

We investigate the power of our retrieval augmentation branch to show the superiority of explicitly introducing extra knowledge via a transfer learning approach. We visualize the feature distribution of Letter-low with t-SNE in Figure \ref{fig:case}, and we focus on a specified class ID-15 in this case study. In (a), we first visualize the feature distribution when ID-15 is a head class, and has the exact same number of samples as other classes. 
We can observe that ID-15 generalizes well and has a clear decision boundary with neighboring classes due to abundant training samples. In (b), we limit the training samples of ID-15, so it becomes a tail class. The feature distribution of ID-15 collapses into a relatively small scope due to the lack of intra-class variability. In (c), we augment the tail class of ID-15 with relevant graphs collected from our retrieval branch. The retrieval features are scattered around the original features of ID-15 in the high-dimensional embedding space, which improve the variety of tail class samples and promote the generalization ability in the tail class.

\section{Conclusion}

This paper presents a retrieval augmented hybrid network (\method{}), which combines transfer learning and decoupling training to boost the learning of both the feature extractor and the classifier simultaneously. \method{} leverages a retrieval augmented branch and a balanced supervised contrastive learning module to jointly learn a balanced feature extractor under the long-tailed setting. Moreover, we re-balancing the classifier weight in norms by applying two weight regularization techniques, which prevent the head classes from dominating the training and improves the tail classes' performance. Extensive empirical studies on six benchmarks show that our approach outperforms competitive baselines on long-tailed data. A potential direction for future work is to enhance the retrieval module by incorporating multi-modal knowledge.

\section*{Acknowledgements}
This paper is partially supported by the China Postdoctoral Science Foundation with Grant No. 2023M730057 as well as the National Natural Science Foundation of China with Grant No. 62276002.

% \end{acks}
\balance
\bibliographystyle{ACM-Reference-Format}
\bibliography{sample-base}

\end{document}